# TOWARDS THE INDUCTIVE ACQUISITION OF TEMPORAL KNOWLEDGE

Kaihu Chen

Artificial Intelligence Laboratory
Department of Computer Science
University of Illinois at Urbana–Champaign

## ABSTRACT

A method is proposed for the inductive acquisition of temporal classification rules from large number of training instances. The method is designed for the incremental acquisition of uncertain rules. Refinement (specialization and generalization) operators for the temporal domain are given, and various heuristics in rule invocation and incremental rule generation are discussed. The method has potential applications in the acquisition of temporal knowledge in many realtime domains.

## 1. Introduction

The acquisition of characteristic descriptions from preclassified examples is an intensively investigated field. Most of the previous researches [Michalski, 1983; Quinlan, 1983; Mitchell, 1977; Rendell, 1985] were limited to the acquisition of timeless, static classification rules. The SPARC system [Michalski & Dietterich, 1985] addressed the problem of discovering rules that qualitatively predict the plausible continuation of a sequence of events in semi–temporal sequential domains, but the program actually falls into the part–to–whole generalization category instead of the instance–to–class generalization category [Michalski, Ko & Chen, 1986] which we wish to investigate here. Developments in temporal logic [Allen, 1983; Vere, 1983] addressed the representation and planning in realtime, but said nothing about inductive acquisition of temporal knowledge.

Here a method is proposed for the incremental acquisition of uncertain class descriptions from preclassified temporally described training instances, called **episodes**. The introduction of temporal knowledge into an inductive learning system demands that the system be able to address the following issues:

- Uncertainty: handling of noisy data is compounded by the introduction of temporal knowledge. It is unrealistic to expect noise–free data from realtime domains; it is imperative that the system be capable of handling noisy data in an appropriate way. Moreover, since perfect class descriptions may not exist or are too costly to find, it is desirable to have the capability to generate "imperfect" rules with variable certainty, and improve these rules until satisfaction.

- Incremental learning: since the acquisition of temporal rules is an expensive process (due to its complexity), incremental rule refinement mechanism [Michalski, 1985] becomes a much more attractive alternative than the batch approach.

The method proposed here is an attempt to address the acquisition of temporal knowledge in the form of classification rules from examples.

## 2. Formulation of the Problem

A training instance, called an **episode**, is given as a list of temporal descriptions of certain objects. Objects are assumed to be uniquely identifiable through time. The following is an example of an episode:

$$<e_1, e_2, e_3, ...> \rightarrow Class_i$$

where an event $e_{ij}$ of an episode is specified as a triplet $\{d_{ij}, ts_{ij}, tf_{ij}\}$, $ts_{ij}$ and $tf_{ij}$ are absolute temporal marks that specify the starting time and finish time of the description $d_{ij}$. The event description $d_{ij}$ is expressed in a representation language powerful enough to describe relationships between objects, such as first order predicate logic or annotated predicate logic [Michalski, 1983]. Given a set of such training instances, the task of an inductive engine is to discover generalized descriptions for each class that have the lowest uncertainty (plus other criteria) using a set of refinement (generalization/specialization) operators.

**The Refinement Operators.** Refinement operators can be divided into two categories: static refinement operators and temporal refinement operators. Static refinement operators are defined on the static language representations. Many static

37

refinement operators, such as "closing an interval", "dropping conjunctive terms", "turning constants into variables", etc., are described in [Michalski, 1983].

Temporal refinement operators are used to derive temporal relationships not explicitly specified in the given data. Temporal generalization from a pair of conjointed events $e_1$ and $e_2$ can be derived by turning the temporal marks into variables, then adding (by conjunction) the relationships between these temporal variables to the expression. For example, given two events

$$e_1 = \{d_1, ts_1, tf_1\}$$
$$e_2 = \{d_2, ts_2, tf_2\}$$

the following temporal attributes can be generated:

$$T_1 = ts_2 - ts_1$$
$$T_2 = ts_2 - tf_1$$
$$T_3 = tf_2 - ts_1$$
$$T_4 = tf_2 - tf_1$$
$$T_5 = tf_1 - ts_1$$
$$T_6 = tf_2 - ts_2$$

The temporal attributes $T_1$ to $T_4$ manifest the temporal relationship between $d_1$ and $d_2$, while $T_5$ and $T_6$ represent the **duration** of $d_1$ and $d_2$ respectively. These attributes can be generalized/specialized like any other linear attributes, where refinement operators, such as closing an interval or climbing hierarchy, can be applied. For example,

$$\{d_1, TS_1, TF_1\} \& \{d_2, TS_2, TF_2\} \& [T_1 = 5] \rightarrow Class_1$$

can be generalized to

$$\{d_1, TS_1, TF_1\} \& \{d_2, TS_2, TF_2\} \& [T_1 > 0] \rightarrow Class_1$$

which says the description for $Class_1$ is "$d_2$ occurs after $d_1$ has finished". Here variables are represented in uppercase, and constants in lowercase letters. This way, a total of six simple generalizations can be produced. Arbitrary combination of these six generalizations also constitute a generalization of the original expression.

Temporal marks, like static structured attributes, can have its own value hierarchy. A set of refinement operators based on this value hierarchy can be created. For example, the temporal mark **12:00 pm** can be generalized to **daytime**, and **January 10** can be generalized to **winter**.

Another temporal generalization operator, similar to the closing-gap operator for linear attributes, works by closing temporal gaps. For example, the rule

$$\{d_1, t_1, t_2\} \& \{d_1, t_3, t_4\} \rightarrow Class_i$$

can be generalized by closing the temporal gap to produce:

$$\{d_1, t_1, t_4\} \rightarrow Class_i$$

assuming that $t_4$ is greater than $t_1$.

## 3. Rule Refinement

Instead of processing training instances and generating classification rules in batch, an incremental learning program must be able to generate new rules by modifying the existing ones in the rulebase. With the set of intermediate (possibly imperfect) classification rules previously learned, the program must make the following decisions when given a new training instance:

- How to select from the rulebase those rules that have the highest promise of being improved by the current training instance. This is the problem of **rule invocation**. Here we essentially assume that the number of rules in the rulebase is so great that exhaustive rule invocation is impractical.

- How to repair the invoked rules in a most efficient way. This is the problem of **rule generation**.

The basic algorithm for incremental rule repairing is shown below:

(1) Input a training instance
(2) Invoke rules from the rulebase.
(3) Refine the invoked rules, then put them back into the rulebase.
(4) If satisfactory rules have been found, then exit; else go to (1).

From the point of view of inductive learning, the introduction of temporal information increase the repertoire of refinement operators available to the inductive engine. Classic representation independent inductive algorithms, such as Mitchell's version space [Mitchell, 1977], are still valid, but their exhaustive nature in rule invocation and rule

38

generation make them too inefficient to handle temporal information. In the following sections heuristics for rule invocation and rule generation are discussed.

**Rule Invocation.** The invocation of rules in the rulebase is decided by their **merits**, which measured the desirability and the prospect of improvement of a given rule. Three types of merit-measuring criteria are distinguished: **efficacy criteria**, **cosmetic criteria**, and **interestingness criteria**. Efficacy criteria measure the confidence level and the performance of a rule, disregarding the complexity in its syntactical form. The confidence level of a rule is proportional to the number of training instances covered by the rule. The performance of a classification rule can be measured by its entropy (defined below). The cosmetic criteria measure desired syntactical properties of a rule such as the simplicity of the form of a rule. The interestingness criteria measure the expected improvement that can be achieved for a given rule.

In the following sections a performance criterion for classification rules based on entropy measurement, and an interestingness criterion based on the **expected maximal entropy reduction** are described. A **uniformness** measure that aids the decision making in rule refinement is also discussed.

**Entropy-Based Rule Invocation.** The purpose of rule refinement is to generate rules that satisfy predefined criteria, such as the simplicity or the generality of the rules. One criterion for measuring the performance of a classification rule can be defined in terms of its **entropy**. The entropy of a classification rule $R$ is given by the following formula:

$$entropy(R) = -p^+ \log p^+ - p^- \log p^-$$

where $p^+/p^-$ are the percentage of positive/negative instances covered by $R$ respectively. A **consistent** classification rule covers either all positive or all negative instances, thus has zero entropy; otherwise it has non-zero entropy. Entropy thus can be viewed as a partial measurement of the certainty of a rule. The **expected improvement** of a classification rule can be defined as the expected maximal reduction in its entropy as a new training instance is registered. Rule invocation based on the EMER (Expected Maximal Entropy Reduction) value of the rules has many interesting properties. It can be shown that EMER corresponds closely to our intuitive notion of the "interestingness" of classification rules. The **interestingness of a training instance** $e$ to a rule $R$ can be defined as the entropy change of $R$ caused by $e$. This measurement provides a way to pinpoint:

- training instances that can be important new evidences in forming new hypotheses (which cause high EMER).
- spurious training instances that can be **noise**, which cause high EMER in rules with **high confidence** (which cover many training instances) in a not-too-noisy domain. These events should be either verified or deleted from the training set.
- commonplace training instances (which cause low EMER) that are forgettable by the system.

Entropy reduction can be viewed as a form of information gain. Rule invocation based on EMER thus maximizes the prospect of gaining information through refinement.

**Uniformness-Based Rule Invocation.** In light of entropy, many inductive algorithms [Michalski, 1983; Quinlan, 1983; Mitchell, 1977; Rendell, 1985] can be viewed as algorithms that generate entropy plateaus in the description space. Programs such as ID3 [Quinlan, 1983] and PLS/1 [Rendell, 1985] explicitly use entropy or entropy-like measures to find maximum entropy gradients in order to divide the space into regions that are relatively flat in entropy. The Aq algorithm [Michalski, 1983] generates zero entropy regions by maximally extending from a zero entropy point (a positive instance). In ID3 or version space [Mitchell, 1983] the need for further specialization is detected by the mere fact that the given rule has non-zero entropy, while in PLS/1 it is detected by a gradient exceeding certain threshold. To counteract the complexity of the temporal domain, we propose a strategy of **lazy evaluation** as an efficient way of detecting the non-uniformness in the entropy distribution of a given rule. A classification rule $R$ is considered **non-uniform** (in entropy) if there exist in the rulebase a specialization of $R$, $R_s$, such that the difference in entropy between $R$ and $R_s$ exceeds certain threshold. A rule is considered to be uniform until proven otherwise by one of its specializations. By evaluating the entropy relationships between related rules in the rulebase, it is possible to prognosticate the internal entropy structure of a rule, hence deciding the proper refinements that should take place.

There are many ways one can read from the uniformness of a rule in order to derive more information:

- A non-uniform rule may indicate over-generalization. In this case further specialization is required.
- A non-uniform rule may indicate that the concept to be learned is intrinsically probabilistic. In this case, further specialization is fruitless. Probabilistic concepts (assuming uniform probability distributions) can be identified as uniform rules with non-zero entropies.
- The non-uniformness can be the result of spurious data, or noise. Such data should be verified, deleted, or the concept can be treated as probabilistic.
- The non-uniformness may be the result of the intrinsically fuzzy boundaries in the concepts to be learned, as is the case in many human concepts. In this case, a region that is relatively flat in entropy can be used to approximate the intended concept.

Rule refinement based on the uniformness measurement will eventually yield rules that are entropy plateaus. Perfectly consistent zero entropy rules can be generated if they do exist, otherwise variable precision rules still can be discovered.

**Rule Generation.** Rule generation can be achieved by either generating totally new rules by applying refinement operators to the old rules, or repairing incrementally the existing rules.

Some rules of thumb for rule refinement are listed below:

(1) If the number of training instances that cause non-uniformness are few, then these instances can be registered as **exceptions** to the rule.

(2) If the number of training instances that cause non-uniformness are moderate, then generate a local generalization of these instances and registered them as **exceptions** to the rule [Reinke & Michalski, 1985].

(3) If no rule repairing seems to work, then generate a completely new rule. This is detailed in the following section.

Since rule repairing by registering exceptions does not requirement reevaluation of the repaired rules, it is a very efficient form of rule refinement and is preferred whenever possible.

**Dependency Directed Rule Generation.** The dependencies between rules (of the same class/action) are marked by the generalization relationship between them. By examining the uniformness and performance of a rule, and the generalization relationship between rules, it is possible to derive powerful heuristics that suggest the proper refinements that should take place. Given two rules $R$ and $R_G$ such a that $R_G$ is a generalization of $R$, derivable by applying the generalization operator $G$ to $R$:

- an over-generalization is indicated if the difference between the entropy of $R_G$ and $R$ is greater than certain threshold. If $R$ is uniform, then a generalization operator $G_s$ that is more specific than $G$ can be used to generate an intermediately general rule $G_s(R)$. This amounts to extending the coverage of a known entropy plateau until an "entropy cliff" is reached. The **extend-against** operation in the Aq algorithm can be used here as an efficient way of reaching the entropy cliff.

- an over-specialization is indicated if the entropy of $R_G$ is the same as $R$, and both $R$ and $R_G$ are uniform. In this case certain static criteria can be used to select the one rule that achieves higher score.

Another type of dependency comes in the form of associations through temporal/spatial proximity. Two rules $R_1$ and $R_2$ can be refined to a common specialization $R_S$ (where $R_S$ is more specialized than both $R_1$ and $R_2$), if they are temporal/spatially associated. Temporal association is marked by the facts that the two rules in question tend to be invoked at the same time (or nearly the same time). Spatial association is marked by the fact that the rules tend to be invoked by the same object (or same set of objects).

## 4. Knowledge Space Decomposition

Another technique to reduce of the complexity of the inductive process is to divide the description space into hierarchies in a fashion similar to those used in INDUCE [Larson, 1977]. In the INDUCE method, the description space is divided into a structure-only space and an attribute-only space. The structure space is searched first to find relationships that characterize the given training instance, then the attribute-only space is searched to fill out the detail generalization. The complexity is thus reduced as a result of reduced search space and simplified graph matching. In the temporal domain,

40

this idea can be extended further by dividing the structure-only space into a static-relation space and a temporal-relation space. The searching precedence can then be arranged in the order of temporal-relation space first, static-relation space second, and attribute-only space last. Such searching precedence is by no means foolproof, thus should be treated as a form of inductive bias that is subject to change from domain to domain.

### 5. Conclusions and Perspective

The application of inductive inference to real-time domain have been hampered by many problems. Notably the presence of uncertainty, the lack of expressive power in the representation languages, the lack of strong heuristics, and the batch nature of many previous systems. As the complexity of the problem domain increases, the capability to learn incrementally becomes increasingly important. The method presented here adopted the empirical approach to learning, thus usually requires large number of trainging examples in order to generate reasonable hypotheses. Analytical learning, where hypotheses are generated through deductive reasoning [Dejong 1986; Mitchell 1985], can also be incorporated into the method by the use of causal models and a deductive element.

### ACKNOWLEDGEMENTS

The author is grateful to Ryszard Michalski and Robert Stepp for their comments and suggestions. This work was supported in part by the Defense Advanced Research Project Agency under grant N00014-K-85-0878, Office of Naval Research under grant N00014-82-K-0186 and National Science Foundation under grant DCR-84-06801.